\documentclass[journal]{IEEEtran}
\usepackage{amsmath,amsfonts}
\usepackage{algorithmic}
\usepackage{algorithm}
\usepackage{array}
\usepackage[caption=false,font=normalsize,labelfont=sf,textfont=sf]{subfig}
\usepackage{textcomp}
\usepackage{stfloats}
\usepackage{url}
\usepackage{verbatim}
\usepackage{graphicx}
\usepackage{cite}
\usepackage[hidelinks]{hyperref}
\usepackage{orcidlink} 
\usepackage{booktabs}
\usepackage{array}
\usepackage{colortbl}
\usepackage{amssymb}% http://ctan.org/pkg/amssymb
\usepackage{pifont}% http://ctan.org/pkg/pifont
\newcommand{\cmark}{\ding{51}}%
\newcommand{\xmark}{\ding{55}}%
\usepackage{multirow} 
\usepackage[table]{xcolor} % For coloring the table rows
\usepackage{enumerate}

\begin{document}

\title{Few-Shot Adaptation Benchmark for Remote Sensing Vision-Language Models}

\author{Karim El Khoury$^{\orcidlink{0000-0002-1594-6531}}$, 
        Maxime Zanella$^{\orcidlink{0009-0009-4030-3704}}$, 
        Christophe De Vleeschouwer$^{\orcidlink{0000-0001-5049-2929}}$, 
        and Benoît Macq$^{\orcidlink{0000-0002-7243-4778}}$
\thanks{K. El Khoury, M. Zanella, C. De Vleeschouwer, and B. Macq are affiliated with UCLouvain, Belgium. M. Zanella is also affiliated with UMons, Belgium.}%
\thanks{M. Zanella is funded by ARIAC (Walloon region grant n\textsuperscript{o}2010235) and C. De Vleeschouwer by the Fonds de la Recherche Scientifique (FNRS).}%
\thanks{Computational resources were provided by LUCIA (Walloon region grant n\textsuperscript{o}1910247) Corresponding author: \texttt{karim.elkhoury@uclouvain.be}.}
}

\maketitle

\begin{abstract}
Remote Sensing Vision-Language Models (RSVLMs) have shown remarkable potential thanks to large-scale pretraining, achieving strong zero-shot performance on various tasks. However, their ability to generalize in low-data regimes, such as few-shot learning, remains insufficiently explored. In this work, we present the first structured benchmark for evaluating few-shot adaptation methods on RSVLMs. We conduct comprehensive experiments across ten remote sensing scene classification datasets, applying five widely used few-shot adaptation strategies to three state-of-the-art RSVLMs with varying backbones. Our findings reveal that models with similar zero-shot performance can exhibit markedly different behavior under few-shot adaptation, with some RSVLMs being inherently more amenable to such adaptation than others. The variability of performance and the absence of a clear winner among existing methods highlight the need for the development of more robust methods for few-shot adaptation tailored to RS. To facilitate future research, we provide a reproducible benchmarking framework and open-source code to systematically evaluate RSVLMs under few-shot conditions. The source code is publicly available on Github: \href{https://github.com/elkhouryk/fewshot_RSVLMs}{https://github.com/elkhouryk/fewshot\_RSVLMs}.
\end{abstract}


\section{Introduction}

Remote Sensing (RS) imagery has become essential in a wide range of applications, ranging from environmental monitoring,
to data-driven farming, as well as emergency disaster response~\cite{ yuan2020deep,phang2023satellite,streamlined2024fast}. These diverse applications require swift and precise scene classification to extract valuable insights from available visual data. Early works in RS scene classification primarily relied on supervised learning methods~\cite{cheng2020remote}. Despite their success, supervised learning methods face significant limitations, such as poor generalization capabilities. Moreover, supervised learning requires extensive carefully labeled datasets for training which is a big issue when considering the large-scale nature of RS imagery. As a result, researchers have shifted focus towards alternative strategies that aim to reduce dependency on extensive annotations and improve the scalability of RS image scene classification models~\cite{miao2022semi,protopapadakis2021stacked,chen2023semiroadexnet,QIU2024368}.

\vspace{3mm}

\par Recently, Vision-Language Models (VLMs) like CLIP~\cite{pmlr-v139-radford21a} overcome these challenges by leveraging large-scale image-text pair datasets for self-supervised contrastive learning, achieving impressive performance across various tasks, including zero-shot image classification with arbitrary class descriptions like \texttt{"a satellite photo of a [class]."}. However, given that foundation models like CLIP are trained on natural images, there is ongoing work to fine-tune these models for RS tasks. These advancements pushed the RS community to develop large image-text datasets and subsequently remote sensing vision-language models (RSVLMs)~\cite{zhang2024rs5m, liu2024remoteclip, wang2024skyscript,pang2024h2rsvlm, Muhtar2024lhrs}, leading to rapid progress in zero-shot scene classification benchmarks~\cite{el2025enhancing}. However, the adaptation ability of RSVLMs under weak supervision, i.e. few-shot setting, is still well under-explored.

\vspace{3mm}

\par Numerous approaches have been considered to improve classification performance based on few labeled examples. Among them, Prompt tuning has emerged as a dominant strategy, refining textual and/or visual tokens to enhance adaptation~\cite{zhou2022learning, khattak2023maple, cocoop, kgcoop, prograd, plot, khattak2023self}. Methods like CoOp~\cite{zhou2022learning} optimize learnable common continuous tokens attached to the class names, described as a context optimization. MaPLe~\cite{khattak2023maple} extends this strategy by introducing learnable visual tokens in addition to textual ones. A second line of research focuses on adapter-based methods, which add a small subset of parameters to improve efficiency~\cite{clip-adapter, tip-adapter, yu2023task, silva2024closer, zanella2025vocabulary}. Tip-Adapter~\cite{tip-adapter} uses a cache-based approach by combining stored feature representations with the original zero-shot prediction. TaskRes~\cite{yu2023task} introduces task-specific residual tuning, adapting the initial text embedding of each class prompt. A third path of investigation considers low-rank fine-tuning, exemplified by CLIP-LoRA~\cite{zanella2024low}, which applies low-rank adaptation within both text and visual encoders. While these few-shot adaptation methods have been extensively explored on natural image classification on the base CLIP model, no work has yet to provide a structured benchmark for few-shot adaptation on RS specific-tasks such as scene classification.

\vspace{3mm}

\noindent \textbf{Contributions} of this work:
\vspace{0.5mm}

\begin{enumerate}[I.] % Roman numerals for enumeration
    \item We introduce the first structured benchmark to evaluate and compare RSVLMs under a few-shot setting; performing extensive experiments across ten RS datasets, three specialized vision-language models, and five state-of-the-art few-shot adaptation techniques.
 
    \item We show that zero-shot performance is not representative of the performance obtained in case of few-shot adaptation in RS tasks and identify which models and adaptation strategies are best suited for few-shot scenarios.

    \item We provide a modular and extensible code-base to help further research in few-shot adaptation of RSVLMs.
\end{enumerate}

\newpage

% Full-width table spanning both columns
\begin{table*}[t]
    \centering
    \footnotesize
    \caption{Comprehensive overview of the 10 remote sensing scene classification datasets used in the few-shot benchmark.}

    \resizebox{\textwidth}{!}{%
    \begin{tabular}{lcccccccccc}
        \toprule
        \multirow{2}{*}{\text{\fontsize{9}{7}\selectfont{\textbf{Dataset Properties}}}} & \textbf{AID} & \textbf{EuroSAT} & \textbf{MLRSNet} & \textbf{OPTIMAL31} & \textbf{PatternNet} & \textbf{RESISC45} & \textbf{RSC11} & \textbf{RSICB128} & \textbf{RSICB256} & \textbf{WHURS19} \\[-0.5mm]
           & \text{\fontsize{6}{7}\selectfont\textit{TGRS'17}} & \text{\fontsize{6}{7}\selectfont\textit{IGRASS'18}} & \text{\fontsize{6}{7}\selectfont\textit{ISPRS'20}} & \text{\fontsize{6}{7}\selectfont\textit{TGRS'19}} & \text{\fontsize{6}{7}\selectfont\textit{ISPRS'18}} & \text{\fontsize{6}{7}\selectfont\textit{Proc.IEEE'17}} & \text{\fontsize{6}{7}\selectfont\textit{JARS'16}} & \text{\fontsize{6}{7}\selectfont\textit{Sensors'20}} & \text{\fontsize{6}{7}\selectfont\textit{Sensors'20}} & \text{\fontsize{6}{7}\selectfont\textit{ISPRS'10}} \\
          \cmidrule(lr){1-11} % Double line separating properties and datasets
        Total Samples          & 10000 & 27000 & 109161 & 1860 & 30400 & 31500 & 1232 & 24747 & 36700 & 1000 \\
        Sample Size (in pixels)           & 600x600 & 64x64 & 256x256 & 256x256 & 256x256 & 256x256 & 512x512 & 128x128 & 256x256 & 600x600 \\
        Total Classes         & 30 & 10 & 46 & 31 & 38 & 45 & 11 & 45 & 35 & 19 \\
        Available Training Samples per Class (avg.)  & 166 & 1350 & 1186 & 30 & 400 & 350 & 56
 & 408 & 353 & 25 \\ 
       Balanced Classes  & \xmark & \cmark & \xmark & \cmark & \cmark & \cmark & \cmark
 & \xmark & \xmark & \cmark \\  
        Fine-grained Class Labels   & \cmark & \xmark & \cmark & \cmark & \cmark & \cmark & \xmark
 & \cmark & \cmark & \cmark \\
        \bottomrule
    \end{tabular}}
    \label{tab:dataset}
\end{table*}

\section{Benchmarking Methodology}

\subsection{Evaluation Datasets}
The few-shot benchmark is composed of ten RS scene classification datasets: AID, EuroSAT, MLRSNet, OPTIMAL31, PatternNet, RESISC45, RSC11, RSICB128, RSICB256, and WHURS19~\cite{xia2017aid,helber2018eurosat,qi2020mlrsnet,wang2018optimal,zhou2018patternnet,cheng2017resisc45,zhao2016rsc11,li2020rsicb,xia2010whurs19}. These datasets cover a wide range of sizes, with sample counts ranging from $\thicksim10^3$ to $\thicksim10^5$ per dataset. They also include varying numbers of classes, from coarse-grained classification with 10 classes to fine-grained scene classification with up to 46 classes. For data splits, all samples from each dataset are grouped then divided into a 50\%/25\%/25\% train/validation/test split using a fixed random seed. It is important to note that none of the selected evaluation datasets were used during the pretraining of the implemented RSVLMs. The properties of all ten benchmark datasets are detailed in Table \ref{tab:dataset}. 

\subsection{Remote Sensing Vision-Language Models}

We implemented three RSVLMs in our benchmark: RemoteCLIP~\cite{liu2024remoteclip}, GeoRSCLIP~\cite{zhang2024rs5m}, SkyCLIP~\cite{wang2024skyscript}, as well as the original CLIP~\cite{pmlr-v139-radford21a}. These three RSVLMs are all fine-tuned variants of the original CLIP, primarily differing in the datasets used during their fine-tuning. By including this variety, we aim to assess how different pretraining data influences performance on RS scene classification tasks. Each model is implemented with their respective visual and textual backbones to generate both image and text embeddings. All four models are tested with RS-specific text-prompt templates of the form \texttt{"a satellite photo of a [class]."}. Each model's characteristics are listed in Table~\ref{tab:RSVLMs}.

 \subsection{Few-Shot Adaptation Methods}

In the few-shot setting, each downstream task is defined by a small \emph{support set} \( \mathcal{S} = \{(x_i, y_i)\}_{i=1}^{C \times K} \), where \( x_i \) denotes an input example and \( y_i \in \{1, \dots, C\} \) is its corresponding class label. The support set contains \( K \) labeled examples for each of the \( C \) classes, resulting in a total of \( C \times K \) training samples. Typical values for \( K \), as adopted in our experiments, are 1, 2, 4, 8, and 16. The objective is to adapt a pretrained model using only this limited support set and then evaluate it on a disjoint \emph{query set} drawn from the same classes. In our benchmark, we assess few-shot adaptation using 5 widely-used methods: CoOp~\cite{zhou2022learning}, MaPLe~\cite{khattak2023maple}, TaskRes~\cite{yu2023task}, Tip-Adapter~\cite{tip-adapter}, and CLIP-LoRA~\cite{zanella2024low}. These popular methods encompass a diverse range of adaptation techniques, including prompt tuning (which introduces learnable tokens as new trainable parameters), adapter-based strategies (which insert additional trainable layers, typically at the output of the model), and low-rank fine-tuning (which incorporates learnable low-rank matrices into the intermediate weights). This diversity of techniques provides valuable insight into which parameters are most relevant to tune. For each method, we adopt the original hyperparameter settings specified in their respective publication.

\begin{table}[t]
    \centering
    \footnotesize
    \caption{Detailed characteristics of the three remote sensing vision-language models used in the few-shot benchmark.}

    \resizebox{\columnwidth}{!}{%    
    \begin{tabular}{lccc}
        \toprule
         \multirow{2}{*}{\text{\fontsize{9}{7}\selectfont{\textbf{Model Characteristics}}}} & \textbf{RemoteCLIP} & \textbf{GeoRSCLIP} & \textbf{SkyCLIP} \\[-0.5mm]
            & \text{\fontsize{6}{7}\selectfont\textit{TGRS'24}} & \text{\fontsize{6}{7}\selectfont\textit{TGRS'24}} & \text{\fontsize{6}{7}\selectfont\textit{AAAI'24}}\\
        \midrule
        Base Model & CLIP & CLIP & CLIP \\[0.5mm]
        \multirow{2}{*}{Vision Encoder Backbones} & RN50, ViT B/32, & ViT-B/32, ViT-L/14, & ViT-B/32, \\[-0.75mm]
        & ViT-L/14 & ViT-H/14 & ViT-L/14 \\[0.5mm]
        Pretraining Dataset & RET-3,SEG-4,DET-10 & RS5M & SkyScript \\[0.5mm]
        Total Pretraining Samples & $\thicksim1.5\times10^5$ & $\thicksim5\times10^6$ & $\thicksim5.2\times10^6$ \\
        \bottomrule
    \end{tabular}}
    \label{tab:RSVLMs}
\end{table}

\section{Experiments}
In this section, we present results obtained using our proposed few-shot RS benchmark and discuss several observations. This analysis showcases the types of insights enabled by systematic evaluation across models, datasets, and adaptation strategies, demonstrating the utility of the benchmark for the RS community.

\subsection{Comparing Remote Sensing Vision-Language Models}

\begin{figure*}[t]
    \centering
    \includegraphics[width=\textwidth]{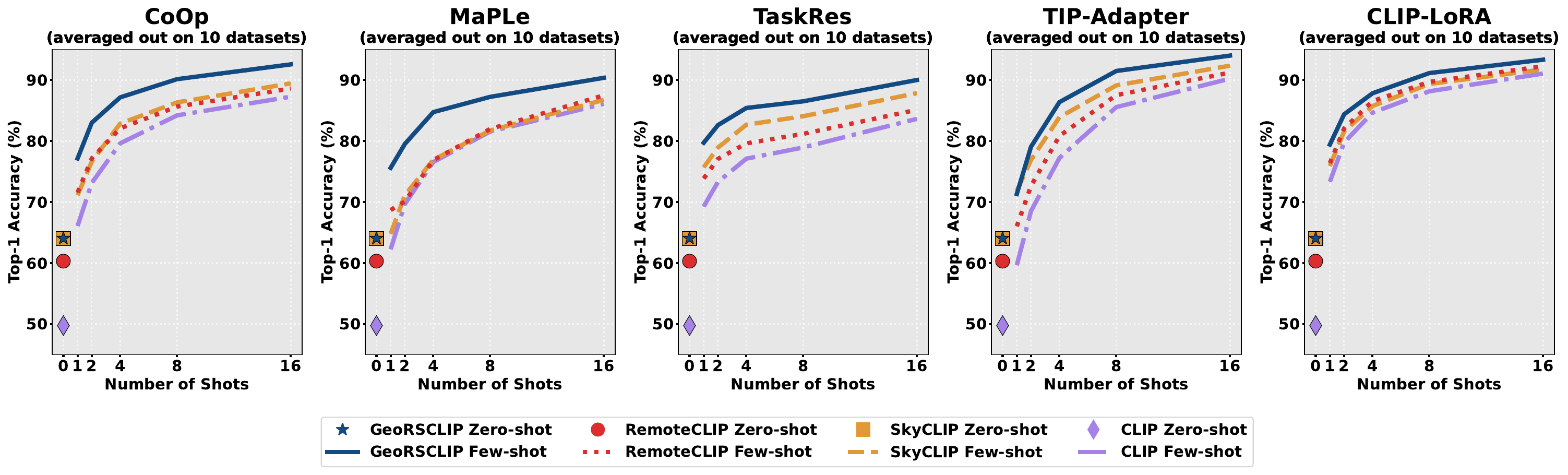}
\caption{
Performance evaluation of four vision-language models (\textbf{\textcolor[HTML]{134A82}{GeoRSCLIP}}, \textbf{\textcolor[HTML]{DC2E2E}{RemoteCLIP}}, \textbf{\textcolor[HTML]{E19838}{SkyCLIP}}, and \textbf{\textcolor[HTML]{A582E7}{CLIP}}) using five different few-shot adaptation methods. Results are computed with ViT-B/32 backbone and represent the average performance across the ten benchmark datasets over three random seeds.
}\vspace{-2mm}
    \label{fig:exp1_ViT_B_32}
\end{figure*}

Figure \ref{fig:exp1_ViT_B_32} presents the average performance of the evaluated models across the ten datasets in our benchmark. All models are compared using the same backbone (ViT-B/32), which allows us to isolate the impact of the pretraining procedure—such as the dataset used, objective functions, and domain alignment—from the effects of architectural differences. The results clearly show that GeoRSCLIP consistently outperforms the other models across all few-shot adaptation methods. An interesting observation is that zero-shot performance is not always a reliable indicator of few-shot adaptation performance. For instance, while GeoRSCLIP and SkyCLIP exhibit similar zero-shot accuracy on average, GeoRSCLIP demonstrates better performance in the few-shot setting. Similarly, although SkyCLIP slightly outperforms RemoteCLIP in the zero-shot setting, their few-shot performance is nearly equivalent across different values of \(K\). Additionally, it is evident that even one-shot adaptation leads to significant improvements over zero-shot evaluation across all four models considered.

% Adjust white space for rules
\aboverulesep = 0.25mm % Default: 0.605mm
\belowrulesep = 0.25mm % Default: 0.984mm

\begin{table*}[t]
\vspace{2mm}
\caption{Detailed results of five different few-shot adaptation methods on GeoRSCLIP ViT-B/32 backbone. Experiments are conducted using three random seeds across the ten benchmark datasets for shot values of 1, 2 and 4. For each scenario, the highest value is marked in \textbf{bold} and the second highest value is \underline{underlined}.}
\label{tab:vitb32_lowshots}
\centering
\setlength\tabcolsep{4pt} % Reduce column padding
\renewcommand{\arraystretch}{1.25} % Adjust row spacing
\vspace{-2mm}
\resizebox{\textwidth}{!}{
\begin{tabular}{llcccccccccccc}
\toprule
\textbf{Shots} & \textbf{Method} & \textbf{AID} & \textbf{EuroSAT} & \textbf{MLRSNet} & \textbf{OPTIMAL31} & \textbf{PatternNet} & \textbf{RESISC45} & \textbf{RSC11} & \textbf{RSICB128} & \textbf{RSICB256} & \textbf{WHURS19} & \textbf{\cellcolor[HTML]{CAD8E4}Average} \\
\midrule
0 & GeoRSCLIP  & 70.9 & 53.2 & 64.9 & 79.6 & 76.1 & 69.1 & 65.9 & 28.8 & 45.1 & 87.0 & \cellcolor[HTML]{CAD8E4}64.1 \\
\midrule
\multirow{5}{*}{1} 
& CoOp & \underline{80.0} & 68.9 & 69.0 & 83.6 & 88.1 & 74.0 & 82.5 & 61.7 & \underline{72.8} & 90.8 & \cellcolor[HTML]{CAD8E4}77.1 \\
& MaPLe & 79.8 & 64.3 & 69.7 & 85.2 & 86.1 & \underline{77.5} & 79.2 & 57.3 & 63.9 & \underline{92.8} & \cellcolor[HTML]{CAD8E4}75.6 \\
& TaskRes & \textbf{81.7} & \underline{71.8} & \textbf{73.7} & \textbf{88.0} & \textbf{89.5} & \textbf{79.9} & \underline{83.7} & \underline{64.9} & 70.5 & \textbf{93.5} & \cellcolor[HTML]{CAD8E4}\textbf{79.7} \\
& Tip-Adapter & 75.0 & 60.1 & 69.9 & 83.0 & 85.6 & 74.7 & 73.8 & 40.7 & 58.8 & 91.5 & \cellcolor[HTML]{CAD8E4}71.3 \\
& CLIP-LoRA & 75.8 & \textbf{73.4} & \underline{70.4} & \underline{87.0} & \underline{89.1} & 75.9 & \textbf{85.0} & \textbf{66.8} & \textbf{77.2} & \textbf{93.5} & \cellcolor[HTML]{CAD8E4}\underline{79.4} \\
\midrule
\multirow{5}{*}{2} 
& CoOp & \textbf{86.1} & 76.0 & \underline{74.7} & 87.2 & \underline{91.3} & 79.7 & \underline{86.6} & \underline{72.5} & \underline{81.5} & 94.5 & \cellcolor[HTML]{CAD8E4}\underline{83.0} \\
& MaPLe & 82.9 & 75.4 & 74.0 & 86.4 & 88.3 & \underline{80.2} & 83.4 & 57.9 & 71.3 & 95.4 & \cellcolor[HTML]{CAD8E4}79.5 \\
& TaskRes  & \underline{84.5} & \underline{76.5} & \textbf{77.9} & \textbf{89.1} & \textbf{91.7} & \textbf{83.6} & 84.7 & 68.5 & 75.6 & 93.7 & \cellcolor[HTML]{CAD8E4}82.6 \\
& Tip-Adapter & 81.3 & 71.5 & 73.0 & 86.2 & 90.0 & 80.0 & 81.9 & 58.1 & 72.6 & \underline{96.1} & \cellcolor[HTML]{CAD8E4}79.1 \\
& CLIP-LoRA & 80.0 & \textbf{83.3} & 73.2 & \underline{88.2} & 90.9 & 79.2 & \textbf{88.1} & \textbf{77.9} & \textbf{86.6} & \textbf{96.3} & \cellcolor[HTML]{CAD8E4}\textbf{84.4} \\
\midrule
\multirow{5}{*}{4} 
& CoOp & \underline{87.9} & \underline{85.7} & 77.6 & 89.0 & 93.0 & 82.3 & \underline{89.1} & \underline{82.7} & \underline{87.8} & 96.4 & \cellcolor[HTML]{CAD8E4}\underline{87.1} \\
& MaPLe & 86.3 & 84.7 & 77.0 & 89.5 & 92.1 & 83.6 & 89.0 & 67.7 & 80.9 & 96.4 & \cellcolor[HTML]{CAD8E4}84.7 \\
& TaskRes & 87.8 & 84.4 & \textbf{79.4} & \textbf{91.4} & 92.8 & \textbf{85.2} & 88.0 & 71.5 & 78.4 & 95.0 & \cellcolor[HTML]{CAD8E4}85.4 \\
& Tip-Adapter & \textbf{89.3} & 80.0 & \underline{77.9} & \underline{90.2} & \underline{93.6} & \underline{84.3} & 88.1 & 77.3 & 85.7 & \textbf{97.2} & \cellcolor[HTML]{CAD8E4}86.4 \\
& CLIP-LoRA & 83.8 & \textbf{88.3} & 76.8 & \underline{90.2} & \textbf{93.9} & 81.8 & \textbf{89.2} & \textbf{85.2} & \textbf{92.5} & \underline{96.5} & \cellcolor[HTML]{CAD8E4}\textbf{87.8} \\
\bottomrule
\end{tabular}}
\end{table*}

\subsection{Comparing Few-Shot Adaptation Methods}
\label{sec:3b}

Table~\ref{tab:vitb32_lowshots} presents detailed results across the ten datasets for the GeoRSCLIP ViT-B/32 model. While CLIP-LoRA performs best on average across the 2- and 4-shot settings, the results vary significantly depending on the dataset and the number of shots. This suggests that the most effective parameters to tune may depend on the specific task. For instance, TaskRes—which modifies output text embeddings—excels on MLRSNet and RESISC45, the datasets with the highest number of classes. On the other hand, CLIP-LoRA—which tunes intermediate model weights—shows strong average performance overall. Additionally, Tip-Adapter tends to underperform in low-shot settings, though the performance gap narrows as the number of shots increases to four. MaPLe performs the worst among the methods studied, likely because it was primarily designed for generalization to new classes and/or domains. These differences underscore the need for more robust and adaptable few-shot adaptation methods tailored to the unique challenges of RS.

\newpage

\subsection{Adaptation Methods Computational Cost}

\begin{figure}[h]
    \centering
    \includegraphics[width=\columnwidth]{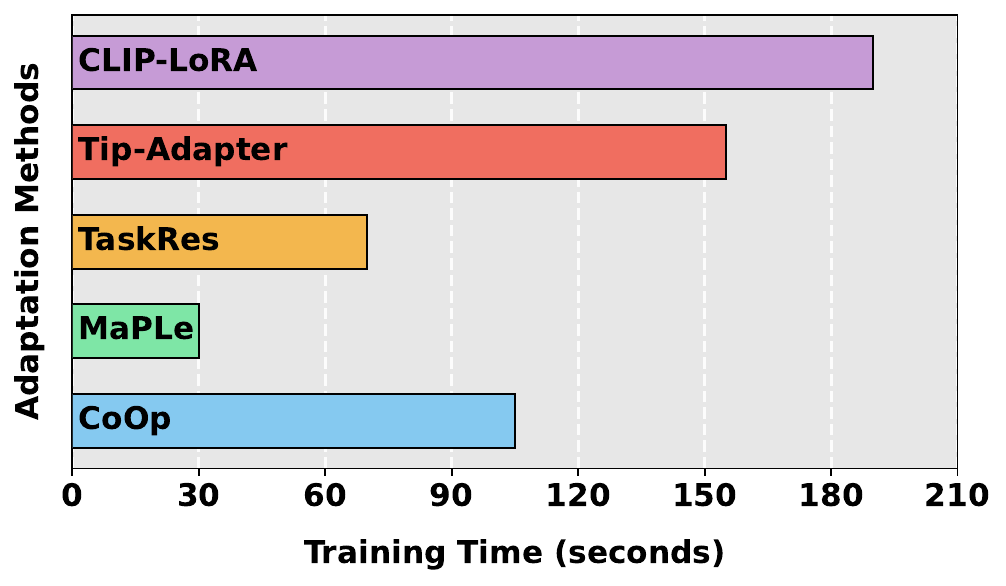}
\caption{Comparison of training time for five few-shot adaptation methods on an NVIDIA A100 80GB GPU, tested on GeoRSCLIP with a ViT-B/32 backbone using 4-shot samples from the MLRSNet dataset (46 classes) and original hyperparameters from their respective publications.}
    \label{fig:cost}
\end{figure}

Figure \ref{fig:cost} highlights the training time for the five few-shot methods considered in the benchmark. While CLIP-LoRA exhibits the best classification performance on average (as shown in Section \ref{sec:3b}), it comes with a compromise in computational cost. The same trade-off holds for CoOp which also necessitates gradient-based optimization over the model. We observe that lighter methods that operate in the embedding space such as TaskRes generally becomes more interesting for achieving a good balance between computational time and accuracy, especially in the 1- and 2-shot settings. We can also note that CLIP-LoRA has the practical advantage that its low-rank matrices can be merged into the model weights after training, incurring no additional inference cost. Given that real-world deployment scenarios may have diverse computational constraints, we show that the most accurate method may not always be the most practical one. We therefore encourage future work to systematically discuss both performance and computational costs when comparing adaptation methods.

\newpage

\subsection{Increasing Training Samples}
Table~\ref{tab:vitb32_highshots} illustrates the impact of increasing the number of training samples to 8 and 16 shots. Interestingly, Tip-Adapter outperforms the other methods, despite being the least effective adaptation method among the five in the 1-shot scenario. Since Tip-Adapter relies on a cache-based approach, while its accuracy improves as the number of shots increases, the longer inference time as shown in Figure~\ref{fig:cost} is to be considered due to the larger size of the stored cache. This highlights the importance of selecting the most suitable adaptation strategy based on the available level of supervision. It also points to promising research directions for developing methods that can maintain strong performance across a broad range of supervision levels.
\vspace{3mm}

\subsection{Scaling to Larger Backbones}
Finally, in Table~\ref{tab:vit_backbone_comparison}, we analyze how the trends observed in earlier sections evolve when scaling to larger backbones. Interestingly, CLIP-LoRA demonstrates superior performance on both the ViT-L/14 (307M parameters) and the larger ViT-H/14 backbone (986M parameters). Furthermore, CoOp, MaPLe, and TaskRes experience significant performance drops when scaled to the ViT-H/14 backbones. The results suggest that low-rank fine-tuning strategies are indeed more robust to model scaling. These results emphasize the need of evaluating performance across different backbone sizes, as the choice of backbone may be influenced by the specific practical constraints of each application.

% Adjust white space for rules
\aboverulesep = 0.25mm % Default: 0.605mm
\belowrulesep = 0.25mm % Default: 0.984mm

\begin{table}[t]
\caption{Average accuracy across the ten benchmark datasets for higher shot values of 8 and 16 using GeoRSCLIP ViT-B/32 backbone. 
Experiments conducted over three random seeds. For each scenario, the highest value is marked in \textbf{bold} and the second highest value is \underline{underlined}.}
\label{tab:vitb32_highshots}
\centering
\setlength\tabcolsep{4pt} % Reduce column padding
\renewcommand{\arraystretch}{1.25} % Adjust row spacing

\resizebox{\columnwidth}{!}{
\begin{tabular}{lccc}
\toprule
\textbf{Method} & \textbf{Average (1-shot)} & \textbf{Average (8-shots)} & \textbf{Average (16-shots)} \\
\midrule
 CoOp & 77.1 & 90.2 & 92.6 \\
 MaPLe & 75.6 & 87.2 & 90.4 \\
TaskRes & \textbf{79.7} & 86.5 & 90.0 \\
Tip-Adapter & 71.3 & \textbf{91.5} & \textbf{94.0} \\
CLIP-LoRA & \underline{79.4} & \underline{91.2} & \underline{93.4} \\
\bottomrule
\end{tabular}}
\end{table}

\aboverulesep = 0.25mm % Default: 0.605mm
\belowrulesep = 0.25mm % Default: 0.984mm

\begin{table}[t]
\vspace{6mm}
\caption{Detailed results of GeoRSCLIP with various backbones. Experiments are conducted using three random seeds across ten benchmark datasets for shot values of 1, 2, and 4. For each scenario, the highest value is marked in \textbf{bold} and the second highest value is \underline{underlined}.}
\vspace{-0.25mm}
\label{tab:vit_backbone_comparison}
\centering
\setlength\tabcolsep{4pt} % Reduce column padding
\renewcommand{\arraystretch}{1.25} % Adjust row spacing

\resizebox{\columnwidth}{!}{
\begin{tabular}{llccc}
\toprule
\textbf{Shots} & \textbf{Method} & \textbf{Average (ViT-B/32)} & \textbf{Average (ViT-L/14)} & \textbf{Average (ViT-H/14)} \\
\midrule
\multirow{5}{*}{1} & CoOp & 77.1 & \underline{81.1} & 73.9 \\
& MaPLe & 75.6 & 78.9 & 67.4 \\
& TaskRes & \textbf{79.7} & 80.3 & 71.9 \\
& TIP-Adapter & 71.3 & 76.9 & \underline{78.9} \\
& CLIP-LoRA & \underline{79.4} & \textbf{83.4} & \textbf{84.3} \\
\midrule
\multirow{5}{*}{2} & CoOp & \underline{83.0} & \underline{86.4} & 80.6 \\
& MaPLe & 79.5 & 82.8 & 69.4 \\
& TaskRes & 82.6 & 83.6 & 76.8 \\
& TIP-Adapter & 79.1 & 82.9 & \underline{83.7} \\
& CLIP-LoRA & \textbf{84.4} & \textbf{87.2} & \textbf{87.1} \\
\midrule
\multirow{5}{*}{4} & CoOp & \underline{87.1} & \underline{89.5} & 85.6 \\
& MaPLe & 84.7 & 85.9 & 74.5 \\
& TaskRes & 85.4 & 83.6 & 77.9 \\
& TIP-Adapter & 86.4 & 89.0 & \underline{89.2} \\
& CLIP-LoRA & \textbf{87.8} & \textbf{90.2} & \textbf{90.4} \\
\bottomrule
\end{tabular}}

\end{table}

\vspace{3mm}
\section{Conclusion}

In this work, we introduced the first structured benchmark for evaluating the few-shot adaptation capabilities of RSVLMs. Our extensive experiments across ten diverse RS scene classification datasets, four VLMs, and five adaptation techniques provide several key insights. We demonstrate that strong zero-shot performance does not necessarily translate to effective few-shot adaptation, emphasizing the need for dedicated evaluation in low-data regimes. Among the models studied, GeoRSCLIP consistently showed superior few-shot generalization, regardless of the adaptation method employed. We also observe that no single adaptation method clearly dominates across all datasets or backbone sizes. While CLIP-LoRA performs well on average, methods like TaskRes, Tip-Adapter or CoOp may be more suitable depending on scenarios-specific constraints. Our benchmark and open-source code provide a reproducible and extensible framework for advancing few-shot learning in RS. As a first step in this direction, future work could investigate few-shot methods explicitly tailored for the unique challenges of RS imagery and explore additional RS tasks beyond scene classification.

\vspace{3mm}

\section*{Code Accessibility}

 Our codebase is designed to easily support the integration of future vision-language models, few-shot learning methods as well as potential integration of supplementary evaluation datasets. The source code in publicly available at: \href{https://github.com/elkhouryk/fewshot_RSVLMs}{https://github.com/elkhouryk/fewshot\_RSVLMs}.

\bibliographystyle{IEEEtran}

\bibliography{refs}
%}

\vfill

\end{document}